\documentclass{sigchi}

\CopyrightYear{2020}
\setcopyright{acmlicensed}
\doi{https://doi.org/10.1145/3313831.XXXXXXX}
\isbn{978-1-4503-6708-0/20/04}
\conferenceinfo{CHI'20,}{April  25--30, 2020, Honolulu, HI, USA}
\acmPrice{\$15.00}

 \toappear{

}

\usepackage{balance}       
\usepackage{graphics}      
\usepackage[T1]{fontenc}   
\usepackage{txfonts}
\usepackage{mathptmx}
\usepackage[pdflang={en-US},pdftex]{hyperref}
\usepackage{color}
\usepackage{booktabs}
\usepackage{textcomp}

\usepackage{microtype}        
\usepackage{ccicons}          

\usepackage{todonotes}

\def\plaintitle{Guru, Partner, or Pencil Sharpener? Understanding Designers' Attitudes Towards Intelligent Creativity Support Tools.}

\def\emptyauthor{}
\def\plainkeywords{Creativity Support Tools; Artificial Intelligence; Machine Learning; Design; Attitudes}

\makeatletter
\def\url@leostyle{%
  \@ifundefined{selectfont}{
    \def\UrlFont{\sf}
  }{
    \def\UrlFont{\small\bf\ttfamily}
  }}
\makeatother
\def\pprw{8.5in}
\def\pprh{11in}

\definecolor{linkColor}{RGB}{6,125,233}
\hypersetup{%
  pdftitle={\plaintitle},
  pdfauthor={\emptyauthor},
  pdfkeywords={\plainkeywords},
  pdfdisplaydoctitle=true, 
  bookmarksnumbered,
  pdfstartview={FitH},
  colorlinks,
  citecolor=black,
  filecolor=black,
  linkcolor=black,
  urlcolor=linkColor,
  breaklinks=true,
  hypertexnames=false
}

\begin{document}

\title{\plaintitle}

\numberofauthors{2}
\author{%
  \alignauthor{Angus Main\\
    \affaddr{Creative Computing Institute}\\
    \affaddr{University of the Arts London}\\
    \email{a.main0920181@arts.ac.uk}}\\
  \alignauthor{Mick Grierson\\
    \affaddr{Creative Computing Institute}\\
    \affaddr{University of the Arts London}\\
    \email{m.grierson@arts.ac.uk}}\\
}

\maketitle

\begin{abstract}
 Creativity Support Tools (CST) aim to enhance human creativity, but the deeply personal and subjective nature of creativity makes the design of universal support tools challenging. Individuals develop personal approaches to creativity, particularly in the context of commercial design where signature styles and techniques are valuable commodities. Artificial Intelligence (AI) and Machine Learning (ML) techniques could provide a means of creating 'intelligent' CST which learn and adapt to personal styles of creativity. Identifying what kind of role such tools could play in the design process requires a better understanding of designers' attitudes towards working with AI, and their willingness to include it in their personal creative process. This paper details the results of a survey of professional designers which indicates a positive and pragmatic attitude towards collaborating with AI tools, and a particular opportunity for incorporating them in the research stages of a design project.
\end{abstract}




\section{Introduction}

The creative industries are an increasingly valuable and strategic part of the global economy \cite{united_nations_creative_2018}. In the UK, the creative sector consists largely of commercial design industries such as product design, interaction design, graphic design, fashion, architecture, and advertising \cite{department_for_digital_culture_media_and_sport_dcms_2018}. This sector is valued at over \pounds100bn, employs over 2 million people, and is growing at nearly twice the rate of the rest of the economy \cite{department_for_digital_culture_media_and_sport_dcms_2019}.

Creativity Support Tools (CST) aim to provide the means of supporting these design industries by enabling "more people to be more creative more of the time"\cite{redondo_creativity_2009}. The design of CST has been well documented in HCI research for over a decade \cite{shneiderman_creativity_2007} \cite{redondo_creativity_2009} \cite{gabriel_creativity_2016} \cite{frich_mapping_2019}. However the wealth of research in this area has not necessarily translated into widely adopted tools for creative workers. 

A key issue at the core of CST design is the complex and subjective nature of creativity itself \cite{boden_creativity_2007}. Definitions and approaches will differ between individuals and industries. Establishing universal methods of support is therefore challenging.

The recent surge in the availability and capability of Artificial Intelligence (AI) and Machine Learning (ML) methods could enable fresh approaches to CST design. A new generation of intelligent and responsive tools could adapt to the specific needs of individual designers, and provide more meaningful and valuable support.

However, a shift from generally passive design tools, towards active, intelligent tools which monitor, learn, suggest, or even collaborate, could have a significant impact on the complex and individualistic creative process. Little data exists on the attitudes of commercial designers towards AI creativity tools, and the role they should play in the creative process. A better understanding of these attitudes could guide the development of intelligent CST, and avoid the design of tools which conflict with the values of designers.

This paper presents a survey of 46 commercial designers, revealing their attitude towards AI and creativity in general, and the role of intelligent CST in particular. It identifies a relationship between the perceived creativity of a design task and the perceived abilities of an AI to support that task. It also indicates specific roles for intelligent CST in the design process.

\section{Background}

\subsection{Understanding Creativity}

To understand designers' attitudes towards creativity it is first necessary to attempt to define what they mean by the term. For a concept so valued and applied within industry, creativity is a complicated concept to define. 

"Human creativity is something of a mystery, not to say a paradox", according to Boden \cite{boden_creativity_2007}, and although the extensive research on the subject ensures it's not entirely mysterious, it's certainly true to say that there are a great many definitions to choose from.

Many of the commonly referenced works on creativity (eg. \cite{boden_creativity_2007} \cite{csikszentmihalyi_creativity:_2013} \cite{sternberg_handbook_1999}) offer interpretations which sometimes compliment, and sometimes contradict each other. Multiple varieties of creativity are suggested in order to account for all facets of the term. For example designers might be 'P-creative' from their own personal point of view, or 'H-creative' from a historical context \cite{boden_creativity_2007}. They could be considered 'G-creative' like a god, or 'N-creative' like nature \cite{still_history_nodate}. Alternatively they could be seen as 'trait creative' where the creativity is a characteristic of their own psychology and behaviours, or 'achievement creative' where it is a quality of their outcomes \cite{eysenck_creativity_1995}. There is no simple method of identifying how an individual designer characterises themselves within this creative landscape.

However, one fairly consistent idea that is present in many definitions of creativity is that it should represent "valuable novelty"\cite{still_history_nodate}. In other words a creative outcome is not only new to a particular domain, but also valuable to someone. But whose values should it satisfy? There will inevitably be some variance in each individual's interpretation of what is valuable, and how that value is achieved.

In the context of commercial design, this subjective view of creativity is reinforced by the intellectual and financial value placed on personal creativity. When individuals are being paid for their creativity, there is an economic advantage to differentiating your 'brand' of creativity from others. This applies at an individual level, but also at a organisational level where companies or studios may foster their own "signature style" \cite{townley_identity_2009}. A signature style can also lead to "identity affirmation" for individuals within larger collaborative design efforts.

When designing CST, the personal creative values of the user need to be identified and navigated. If the complexity of the term means that it's not possible to support designers to create outcomes which objectively meet universally agreed, formal definitions of creativity, then the focus must be narrowed to a specific context of design. 

The challenge then becomes creating tools which support the variety of personal approaches to creativity, and can adapt to the values of particular designers, or design teams. This issue is addressed in the emerging area of intelligent CST, which utilise AI in order to augment or enhance personal creativity.

\subsection{Intelligent CSTs}

In the conclusions of their survey of creativity support systems, Gabriel et al \cite{gabriel_creativity_2016} note the problems associated with designing tools which support creativity across all phases and settings. They conclude that to address this CSTs would need to offer "advanced functionalities, such as adaptation of the system to the behaviour and cognitive patterns of individuals, [which] implies the introduction of artificial intelligence into the creativity support." Their survey highlights that the use of AI to create intelligent CSTs has not been common practice in the many projects they review.

Machine Learning can be used to analyse behavioural or environmental patterns and adapt the support provided as appropriate. This is a potential new form of functionality for CST and an emerging area of research (for example, Gon\c{c}alves et al \cite{madureira_emerging_2017}).  As Gabriel et al suggest, such systems may also be able to better support creativity across "different collaboration settings", by revealing differences in the creative approaches of individual team members and facilitating mediation between them.

However, the introduction of intelligent CST would change the dynamic between tool and user, and perhaps raise the question of not just what functionality the tool offers, but what role it plays in a participatory design process. For example, Lubart \cite{lubart_how_2005} anticipates CSTs could assume the roles of nanny, pen-pal, coach, or colleague. The last category has already been explored by CST projects which create close feedback loops between human creatives, and AI systems, facilitating a design process where human and computer are co-creating an outcome (for example Deterding et al \cite{deterding_mixed-initiative_2017}). By contrast, the kind of ambient intelligent support suggested by Gon\c{c}alves et al most resembles Lubart's nanny role, or perhaps an entirely separate category more akin to an attentive assistant.

Feeding into the question of role or character are the broader cultural perceptions of intelligent assistant tools. These perceptions are likely to have evolved with the popularisation of virtual and voice-based assistants such as Siri, Alexa, Cortana, and Google Assistant, which have gone some way to normalising the idea of personified computational support. Understanding how this kind of support is perceived, and what virtual roles designers are likely to accept in their creative process, is important in developing future CSTs, and a factor addressed in the survey presented here.

In addition, the attitudes of designers may be influenced by ongoing discussion in the press and media about the potential of AI acting creatively in it's own right (\cite{lawrie_could_2019} \cite{sautoy_true_nodate} \cite{baraniuk_artificially_nodate}  \cite{mulhall_bbc_2019}). Whatever the philosophical and technical merits of this argument, the prevalent idea of AI as a creative peer could influence attitudes for better or worse. This research does not take a position on whether or not an AI support tool can act creatively, but instead asks the more fundamental question of what kind of role (creative or otherwise) designers want an AI tool to play in their creative process.

Another important consideration addressed by the survey is the potential attitudinal impact of intelligent CSTs. Given the personal nature of creative design, and the value of signature styles, ascribing CSTs a greater sense of agency in the creative process may introduce issues of ownership. Not necessarily in a legal sense (that may be an issue that needs resolving by a separate domain of research), but ownership in an emotional or intellectual sense. For example, research relating to a poetry CST 'Metaphoria' (Gero \& Chilton, 2019 \cite{gero_metaphoria:_2019}) highlighted that a user's sense of ownership over their creative outcomes was negatively impacted by the use of the tool. This response varied depending on whether the user approached the CST as a "co-creative partner" or "cognitive offloading tool". 

\begin{figure*}
  \centering
  \includegraphics[width=2\columnwidth]{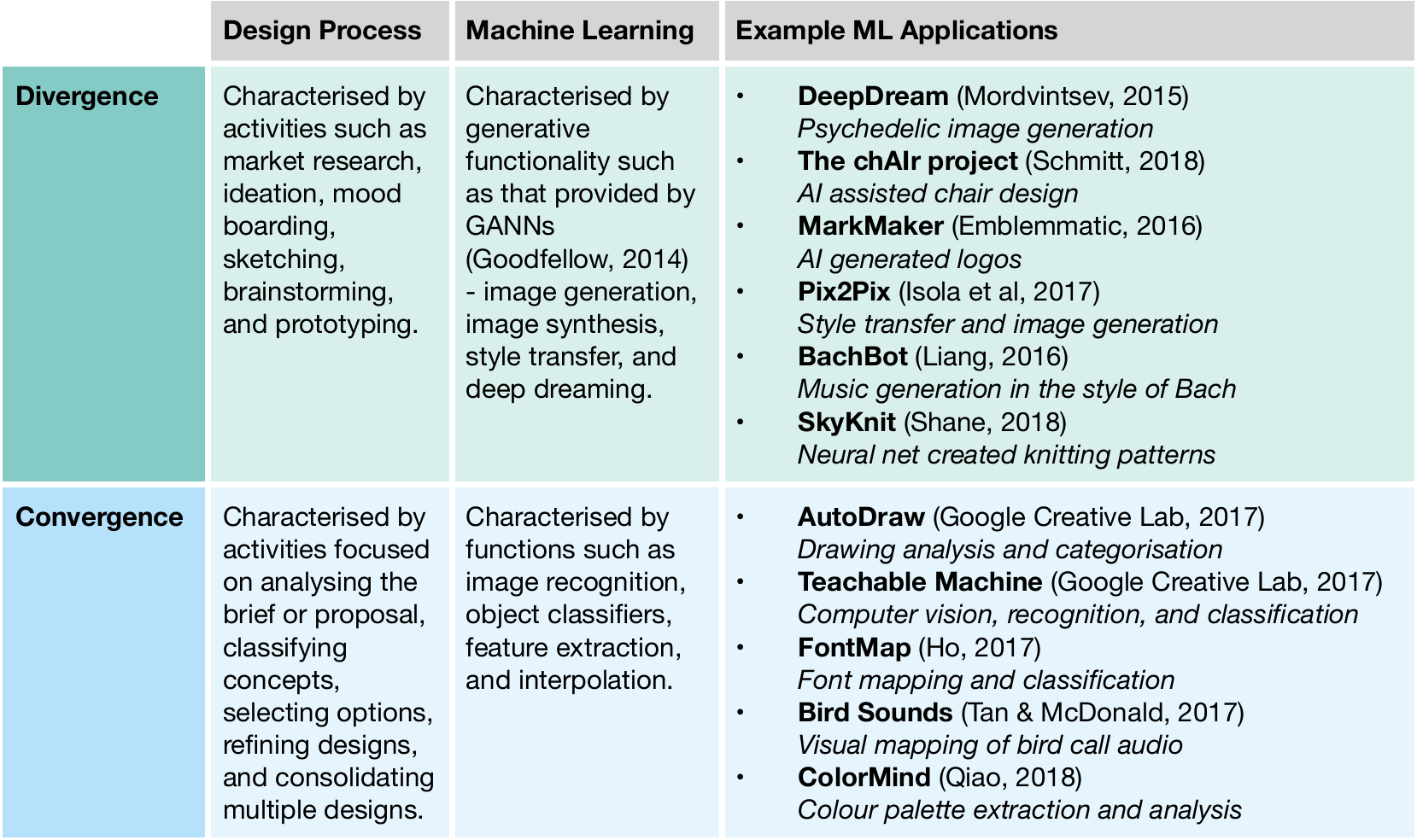}
  \caption{Examples of Divergence and Convergence within the design process and within Machine Learning applications  }~\label{fig:figure1}
\end{figure*}

Despite the technical possibilities of augmenting human creativity with AI, it's possible that emotive issues such as 'sense of ownership' contribute to designers wholly rejecting the idea of intelligent CST playing an active role in their creative outcomes. It should not be ruled out that the most helpful way of allowing a designer to be creative might be to support them in their non-creative tasks (for example answering emails, ordering materials, preparing invoices) or supporting the use of existing creative tools (for example managing files or documenting ideation). These sort of tasks do not require the CST to engage directly in creative work so their role might be considered more similar to an answer machine or pencil sharpener rather than colleague or coach. However, this kind of indirect creativity support might allow designers to spend more time being independently creative. 

Schneiderman \cite{shneiderman_creativity_2007} notes the evolution of computing from supporting productivity to supporting creativity. However, the two are interconnected. It might be that for commercial designers, productivity tools still play an important role in supporting their creative process.

The survey examines these issues by querying whether designers perceive specific tasks within the design process as requiring high or low creativity, and comparing this with the their perceptions of the capabilities of AI to support the task.

\subsection{Aligning ML Functionality With The Design Process.}

Individual approaches to creativity may vary, but within the scope of an organisation or industry there is usually consensus on how the creative process is organised. A basic and commonly used structure is described by the British Design Council's Double Diamond \cite{design_council_uk_eleven_2007}. Consisting of four phases of alternating 'divergent' and 'convergent' activities, the model was based on the design practices of different organisations (for example Alessi, BT, Microsoft, Sony \cite{design_council_uk_eleven_2007-1}). The Double Diamond method has been used in different industries as a basis for structuring the creative process, such as Google's 'Design Sprint' model \cite{google_share_2019}. The iterative nature of the model allows it to be integrated with the Agile methods used by digital design and software companies. 

The model provides a context for understanding when and how creativity is used on a design project, and therefore identifying occasions when the use of CST might be appropriate. The broad structure - Discover, Define, Develop, Deliver - has therefore been used as a means of structuring enquiries for this research.

The Double Diamond separates the design process into broadly divergent and convergent activities, where divergence describes broadening thinking, opening up options, and discovering opportunities, and convergence relates to focusing thinking, selecting options, and finding consensus.

\begin{figure*}
  \centering
  \includegraphics[width=2\columnwidth]{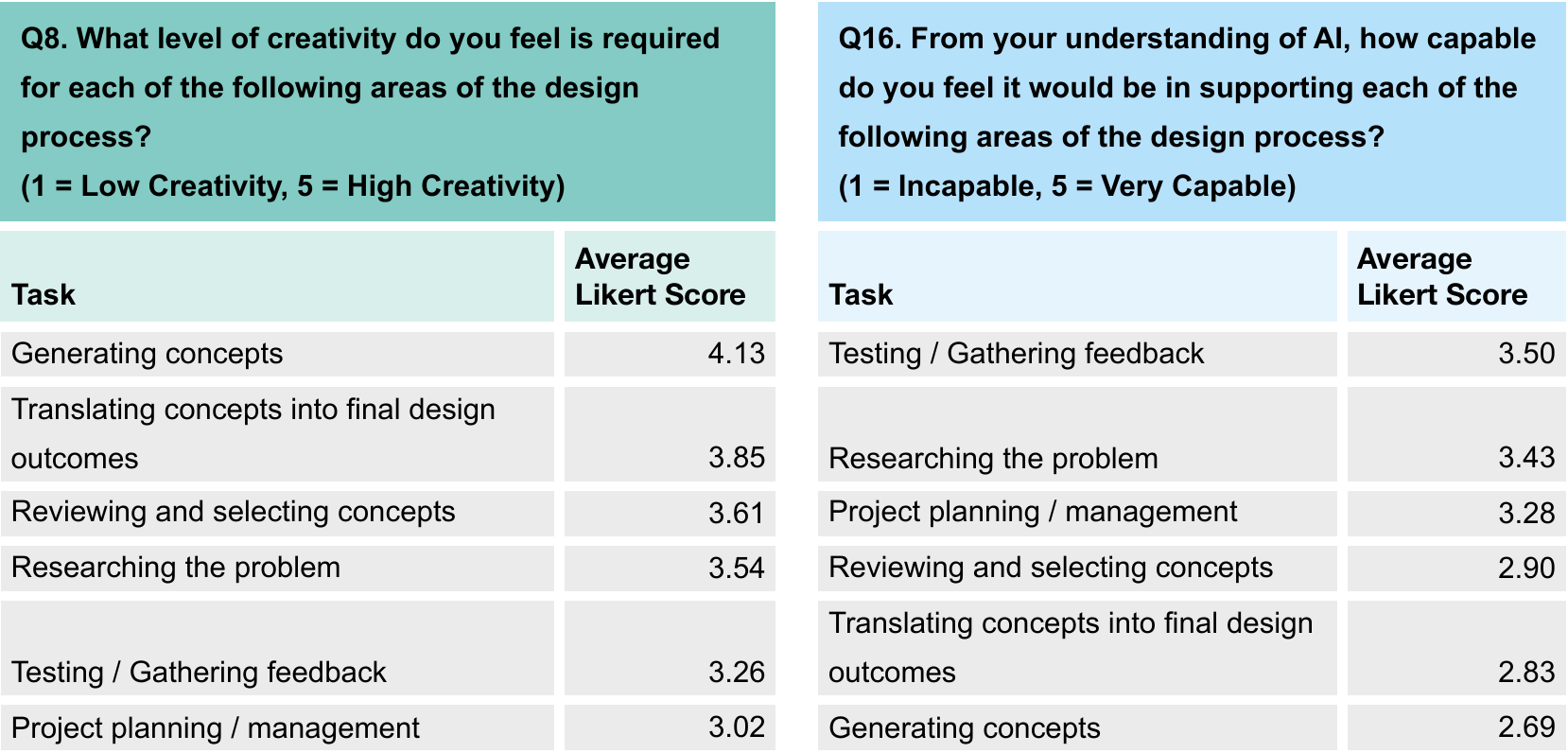}
  \caption{A comparison of results for Q8 and Q16. The Likert scores for each task were averaged and ranked in order of creativity (Q8) and perceived capability of AI to support the task (Q16). }~\label{fig:figure2}
\end{figure*}

There are some parallels between these two categories of activity, and the functionality demonstrated in current implementations of ML (Fig. 1). The survey has been designed to test whether this alignment could be used to direct the functionality of future intelligent CST.

\section{Method}

To test the attitudes of commercial designers towards AI tools and creativity, a survey was conducted of 46 designers from professions that form part of the creative industries \cite{department_for_digital_culture_media_and_sport_creative_2001}. Survey participants were recruited primarily from alumni and students of postgraduate design courses in the UK. 

The survey was distributed as an online form, and consisted of 19 questions split into four separate sections: Demographic Information, Creativity, AI, and Creativity Support. Question topics were aligned between the different sections to test where attitudes towards different concepts might be connected. The attitudes of respondents was measured using 1-5 Likert style questions.

The survey was designed to reveal insights into the following specific areas:
\subsection{What are the defining qualities of creativity for designers?} (Q5) - Respondents were asked to identify terms which they associated with creativity in order to test whether there was a common understanding of the topic of the survey, and to identify key qualities which could direct future CST development.
\subsection{How does the perceived creativity of a design task compare with the perceived ability of an AI to support it?} (Q8 \& Q16) - Two linked questions in separate sections of the survey asked respondents to rate the level of creativity required for specific tasks in the design process, and later asked them to rate their perception of the capability of AI tools to support the same list of tasks. Comparing the responses to these questions was intended to reveal attitudes towards the ability of AI tools to support creative vs. non-creative tasks.

\subsection{How would the use of AI tools effect designers' sense of ownership over creative outcomes?}(Q18) - Respondents were asked how their sense of ownership over a creative outcome would be effected by the use of an AI support tool. This was designed to reveal whether personal emotional attitudes to creativity were likely to mean designers rejected the support of intelligent CST, or whether they felt they needed to modify suggestions in order to gain a sense of ownership.

\subsection{What are designers' general attitudes towards AI technology?} (Q9, Q10, Q11, Q12) - A series of questions asked respondents to identify terms they associated with AI in order to reveal general sentiment towards AI technology, and also asked respondents to rate what impact they felt AI technology would have on their industry in the future. The result of these questions could be used to reveal whether designers attitudes to intelligent CST are consistent with a broader attitude towards AI.

\subsection{What are common barriers to creativity?} (Q7) - Respondents were asked to identify common issues which prevented them from achieving creativity. This was intended to reveal common experiences of the creative process, and to indicate what manner of support a CST could usefully provide.

The results of the survey were collated and analysed using a mixture of parametric and non-parametric methods. 

\section{Results}

The responses provided the following insights into the defined areas of enquiry:

\subsection{What are the defining qualities of creativity for designers?}

\textit{\\Novelty is the most important quality of creativity to designers}

Respondents identified novelty as being the most important quality of a creative design, with over 71\% indicating it was of high, or very high importance.

This corresponds with the common definition of creativity as "valuable novelty". However, the second most highly rated quality did not relate to purpose or utility, but instead ingenuity, which was defined as "demonstrating clever or complex problem solving". As problem solving is integral to design practice, this suggests that designers view creativity as synonymous with good design.

\subsection{How does the perceived creativity of a design task compare with the perceived ability of an AI to support it?}

\textit{\\The design tasks perceived as most creative are the ones perceived as least suitable for an AI to support}

Comparing the answers to questions 8 and 16 ("What level of creativity do you feel is required for each of the following areas of the design process?" and "From your understanding of AI, how capable do you feel it would be in supporting each of the following areas of the design process?") reveals a significant inverse correlation between the perceived creativity of a task, and the perceived ability of an AI to support it.

Analysing the results first through a simple averaging of the Likert scores (fig. 2) shows that the three tasks collectively ranked most creative (generating concepts, translating concepts into final designs, reviewing and selecting concepts), were also the three ranked bottom in terms of AI capability.
\begin{figure}
\centering
  \includegraphics[width=0.9\columnwidth]{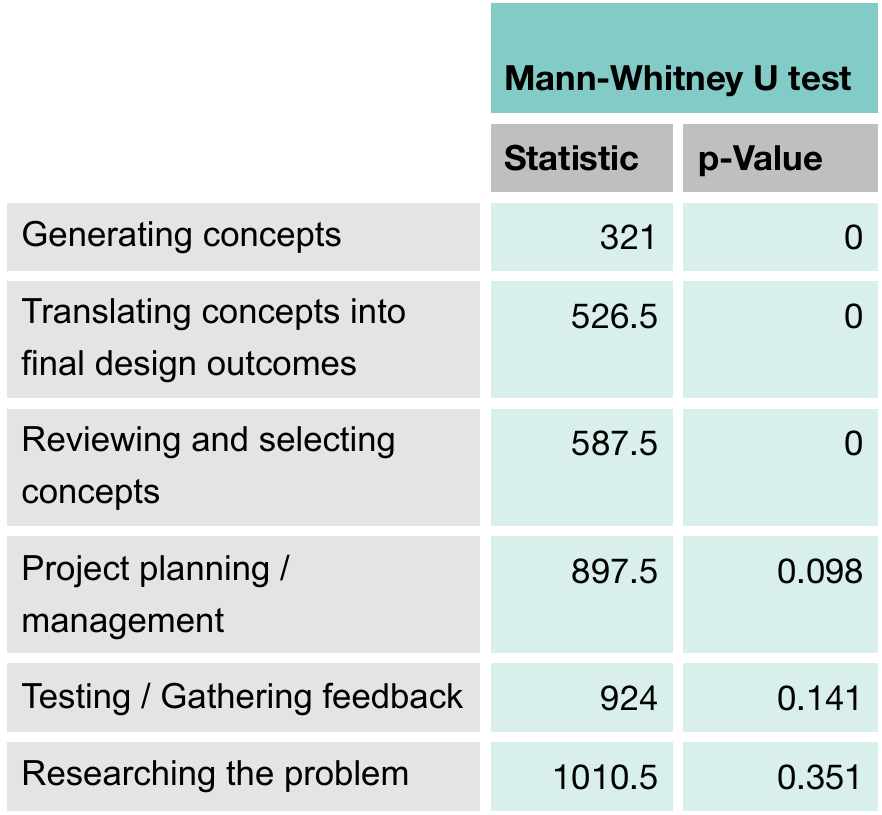}
  \caption{Results of a Mann-Whitney U test applied to the responses to Q8 and Q16 }~\label{fig:figure3}
\end{figure}

Reviewing the data on an individual basis shows that in the majority of cases (71\%), where a respondent indicated a high level of confidence in an AI being able to perform a particular task, it was for a task that they felt required low creativity.

Testing the results using non-parametric methods reveals the tasks where there is the largest difference between perceived creativity of a task, and perceived ability of AI to support it.
A Mann-Whitney U Test and a Wilcoxon Signed-Rank Test were carried out on the results of the two questions to test the distribution of answers. Both tests produced similar outcomes, showing that the largest difference between perceived creativity and perceived capability of AI was for the tasks "generating concepts", "translating concepts into final design outcomes" and "reviewing and selecting concepts". These tasks all had different distributions, indicating that they were unlikely to be rated highly for both creativity and perceived capability of AI.

The tests also showed that the tasks "project planning/management", "testing/gathering feedback" and "researching the problem" had similar distribution of results, indicating that respondents rated the required creativity and AI capability of these tasks more evenly. 

\begin{figure}
\centering
  \includegraphics[width=0.9\columnwidth]{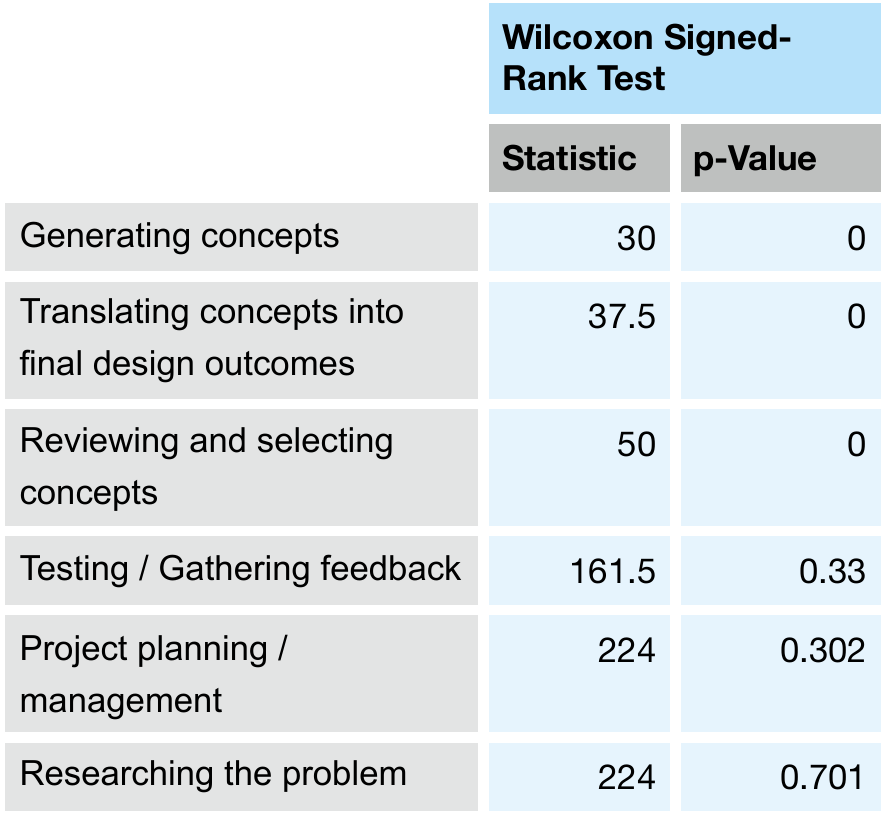}
  \caption{Results of a Wilcoxon Signed- Rank Test applied to the responses to Q8 and Q16 }~\label{fig:figure4}
\end{figure}

These results are also supported by the answers given to question 17, which provided respondents with a longer list of design tasks and asked them to indicate how willing they would be to share the task with an AI. The results showed that the research related tasks "researching existing design solutions" and "researching materials/tools/processes" were (along with "testing") those which they'd be most willing to have performed in part by an AI.

Taking all these results together indicates the design tasks which have the least and most potential to be supported by intelligent CST. At one end of scale, with the least potential, is "generating concepts" which designers are likely to rate as a highly creative task, but unlikely to believe an AI is capable of supporting. At the other end, with high potential is "researching the problem", which designers are likely to rate as requiring at least medium creativity, which is then matched by their confidence in the capability of AI to support the task.

\subsection{How would the use of AI tools effect designers sense of ownership over creative outcomes?}

\textit{\\Designers do not perceive an issue with ownership }

Addressing the issue of perceived ownership of AI supported work raised in Gero \& Chilton \cite{gero_metaphoria:_2019}, Q17 asked "If you used an AI tool to support your creative process, how would it effect your sense of ownership over the outcome?".

The responses indicated that amongst designers there does not seem to be a problem with the perceived ownership of AI supported work, and that it seems unlikely that support would be rejected out of hand.

\begin{figure}
\centering
  \includegraphics[width=0.9\columnwidth]{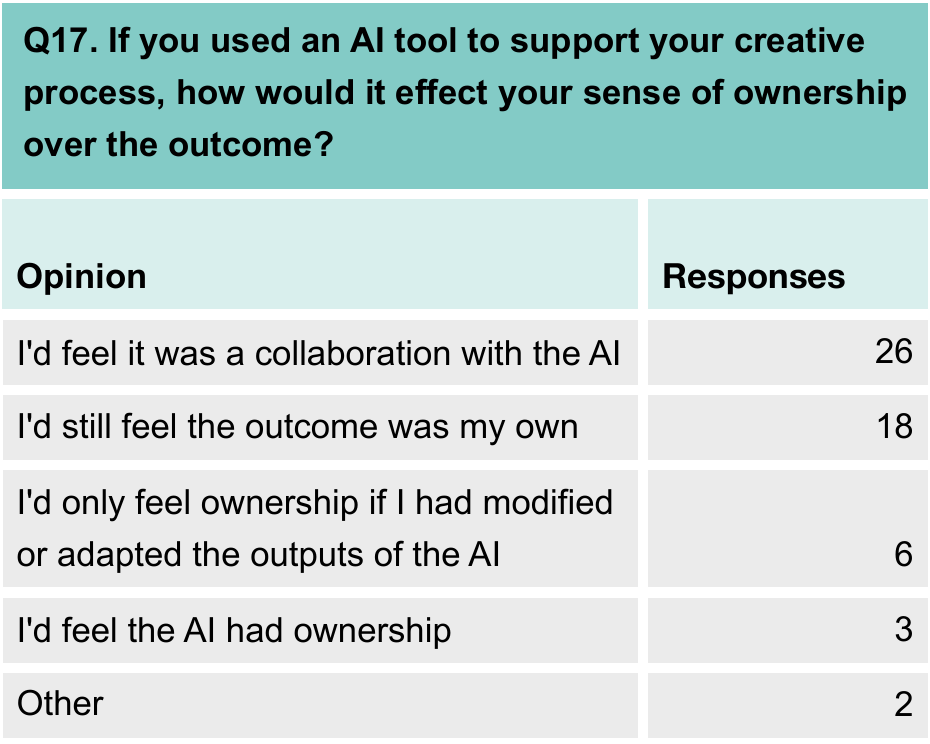}
  \caption{Results for Q17 showing the perceived impact of AI tools on designer's sense of ownership}~\label{fig:figure5}
\end{figure}

Over half the respondents felt that if an AI tool had supported their creativity, they would view the resultant outcome as a collaboration with the AI. This suggests a positive view of the abilities of AI technology, and an inclusive view of its role. 40\% of those surveyed felt they would still have ownership of the outcome. Only a small minority felt that ownership would belong to the AI tool, or that they would need to alter the outputs in order feel a sense of ownership.
\subsection{What are designers general attitudes towards AI technology?}

\textit{\\Designers have a generally positive outlook on AI}

The survey did not reveal any significant negative sentiment relating to AI technology. Most of the respondents (68.9\%) predicted that AI technology would have a high or very high impact on their work. However, this does not seem to be a cause for concern as most of those surveyed (86.6\%) were either optimistic or neutral about the nature of this impact. 

Neither did respondents feel they had a poor understanding of AI technology. 82.2\% felt their understanding was either average or good. Further testing would be needed to determine whether this perception of their own knowledge was accurate. 

\subsection{What are common barriers to creativity?}

\textit{\\Distraction and Fixation are common obstacles to creativity}

The most common obstacles to personal creativity were ranked as "distraction from non-creative tasks" and "too much fixation on task", each being selected by 40\% of respondents. 

The next most identified barriers were "Lack of interest in the problem" (35\%), "Lack of understanding of the problem" (33\%), and "Lack of inspiration" (33\%).

\section{Discussion} 

\subsection{Research over concepts}

The results of the survey indicate that in the context of designers' perceptions of AI technology and creativity, language is important. It's notable that tasks which included the word "concept" (e.g. "generating concepts") were perceived as requiring the highest creativity, and not considered to be within the capabilities of AI. The impression is perhaps that conceptual work is still the domain of human creativity, and it would not be easy for computational intelligence to play a role in this part of design. 

On the other hand, tasks that contained the word "research" hold an interesting middle ground. They were perceived as creative tasks, but there was also confidence in the abilities of AI to perform or support them. These research tasks, relating largely to the Discovery phase of the design process, therefore may have the most potential for future intelligent CST development.

Further research would be needed to understand how much of this perception directly relates to the language used. For example, if words such as "concept" were avoided in the description of task, would it effect designers attitudes towards it? The use of language may be an important factor in the description and positioning of intelligent CST.

The perception of research tasks is also worth considering in relation to the significance of 'novelty' in designers' definition of creativity. If a CST is able to support designers with discovery phase activities such as understanding the problem space of the brief, helping map existing solutions, or diversifying their sources of inspiration, then it could make it easier for them to achieve novelty in their outcomes. 

\subsection{Sense of Ownership}

Despite there being low confidence in the current abilities of AI to play a role in highly creative tasks, there is evidence that designers do have a positive and pragmatic approach to to the role AI tools could play as collaborators in design projects. This may prove significant as the capabilities (and perceived capabilities) of AI improve in the future.  

The willingness to perceive an AI tool as a collaborator provides an alternative view to previous research suggesting the sense of ownership of creative tasks may be negatively impacted by AI tools. This could be due to the particular requirements and attitudes of commercial designers, as opposed to other creatives such as poets or fine artists. This view was expressed in the free text response of one respondent who disagreed that sense of ownership would be an issue as "design is in general a much more collaborative process than Art". Whether that is true or not (many art disciplines require extensive collaboration), designers' perception of the importance of collaboration may be a factor in their use of intelligent CST.

Further research would be needed to gain more understanding of this issue. What this survey does seem to indicate though, is that there is potential for intelligent CST to be used within the creative design process without being seen as intrinsically problematic by most designers.

\subsection{Barriers to Creativity}

The fact that two slightly contradictory issues, Distraction and Fixation, were frequently identified together as barriers to creativity, perhaps underscores the complexity in defining creativity and creativity support opportunities.

As common obstacles however, it could be valuable to address these issues in future CST design. The nature of both obstacles suggests that a support tool fulfilling Lubart's definition of a 'nanny' role \cite{lubart_how_2005} could be most suited to guiding the performance of the designer, and assisting them in avoiding either fixation or distraction from creative tasks. Ambient and adaptive intelligent support tools such as Gon\c{c}alves et al \cite{madureira_emerging_2017} might be best placed to provide this kind of assistance.

The issue of being distracted by non-creative tasks also suggests there is still a role for productivity tools which support creativity indirectly by assisting with non-creative tasks on behalf of designers.

\subsection{Designers' Perception of AI}
More broadly, the survey gives some insight into designers' perception and understanding of AI. This has wider implications at a time when AI functionality is becoming an increasingly common material \cite{holmquist_intelligence_2017} in design outcomes. User experience, interaction, and product designers, for example, are increasingly likely to need to integrate AI into their designs. Some studies have noted the need for better knowledge amongst designers of the capabilities of AI \cite{yang_investigating_2018} \cite{loi_co-designing_2019}, although this survey gives some indication that they already feel they have a reasonable understanding. If designers are open to including AI tools in their creative process in a collaborative role, then this close engagement with AI might lead to better sensitisation \cite{yang_mapping_2018} of designers to the capabilities and limitations of the technology.

Helping designers design \textit{for} AI, by allowing them to design \textit{with} AI, is another emerging area of CST research (eg. \cite{malsattar_designing_2019}) which could be influenced by the findings of this survey.

\section{Conclusion}

Although the scale of the survey is limited, the expertise of the respondents means that the results could reflect the broader sentiments of professional designers. Further research focused on the insights gained from this survey could establish best practice within the design of intelligent CST.

From this initial research however it is possible to point towards potential directions for the development of intelligent CST. In particular it seems that tasks within the Discovery stage of a design project, e.g. inspiration research, creating mood boards, materials research and market analysis, could currently provide the biggest opportunity for intelligent CST. The perceived capability of AI to support these tasks is a useful counterpoint to the general tendency of surveyed designers to associate AI support with non-creative tasks. 

Although designers appear less eager to share conceptual creative tasks such as ideation with AI tools, the survey indicates that they do have a pragmatic approach to collaborating with intelligent tools, and would not necessary feel their ownership of the creative outcome had been compromised as a result. While the abilities of AI tools to support creative work remain unproven in the minds of designers, creating successful intelligent CST will require a careful balancing of these considerations. Although there may need to be a further shifting of perceptions before designers are ready to believe AI tools could fulfil a role of co-creator, it does seem they might accept them in supportive roles such as assistant, collaborator, researcher or facilitator. In other words, positions which reinforce, rather than diminish, their own role as a creative.

%
%
%
%
%

\balance{}

\bibliographystyle{SIGCHI-Reference-Format}
\bibliography{chi}


\begin{thebibliography}{00}


\ifx \showCODEN    \undefined \def \showCODEN     #1{\unskip}     \fi
\ifx \showDOI      \undefined \def \showDOI       #1{{\tt DOI:}\penalty0{#1}\ }
  \fi
\ifx \showISBNx    \undefined \def \showISBNx     #1{\unskip}     \fi
\ifx \showISBNxiii \undefined \def \showISBNxiii  #1{\unskip}     \fi
\ifx \showISSN     \undefined \def \showISSN      #1{\unskip}     \fi
\ifx \showLCCN     \undefined \def \showLCCN      #1{\unskip}     \fi
\ifx \shownote     \undefined \def \shownote      #1{#1}          \fi
\ifx \showarticletitle \undefined \def \showarticletitle #1{#1}   \fi
\ifx \showURL      \undefined \def \showURL       #1{#1}          \fi

\bibitem{mulhall_bbc_2019}
 2019.
\newblock {BBC} {Radio} 3 - {Free} {Thinking}, {AI} and creativity: what makes
  us human?
\newblock   (June 2019).
\newblock
\showURL{%
\url{https://www.bbc.co.uk/programmes/m0005nml}}


\bibitem{baraniuk_artificially_nodate}
{Chris Baraniuk}. 2017.
\newblock Artificially intelligent painters invent new styles of art.
\newblock   (2017).
\newblock
\showURL{%
\url{https://www.newscientist.com/article/2139184/}}


\bibitem{boden_creativity_2007}
{Margaret~A. Boden}. 2007.
\newblock \showarticletitle{Creativity in a nutshell}.
\newblock {\em Think\/} {5}, 15 (2007), 83--96.
\newblock
\showISSN{1477-1756, 1755-1196}
\showDOI{%
\url{http://dx.doi.org/10.1017/S147717560000230X}}


\bibitem{csikszentmihalyi_creativity:_2013}
{Mihaly Csikszentmihalyi}. 2013.
\newblock {\em Creativity: the psychology of discovery and invention\/} (1. ed
  ed.).
\newblock Harperperennial, New York.
\newblock
\showISBNx{978-0-06-228325-2}
\newblock
\shownote{OCLC: 825754060.}


\bibitem{department_for_digital_culture_media_and_sport_creative_2001}
{{Department for Digital, Culture, Media and Sport}}. 2001.
\newblock Creative {Industries} {Mapping} {Documents} 2001.
\newblock   (2001).
\newblock
\showURL{%
\url{https://www.gov.uk/government/publications/creative-industries-mapping-documents-2001}}


\bibitem{design_council_uk_eleven_2007}
{{Design Council UK}}. 2007a.
\newblock {\em Eleven lessons: managing design in eleven global brands. {A}
  study of the design process}.
\newblock {T}echnical {R}eport. London, UK.
\newblock


\bibitem{design_council_uk_eleven_2007-1}
{{Design Council UK}}. 2007b.
\newblock {\em Eleven lessons: managing design in eleven global companies.
  {Desk} research report}.
\newblock {T}echnical {R}eport. London, UK.
\newblock
\showURL{%
\url{https://www.designcouncil.org.uk/sites/default/files/asset/document/ElevenLessons_DeskResearchReport_0.pdf}}


\bibitem{deterding_mixed-initiative_2017}
{Sebastian Deterding}, {Jonathan Hook}, {Rebecca Fiebrink}, {Marco Gillies},
  {Jeremy Gow}, {Memo Akten}, {Gillian Smith}, {Antonios Liapis}, {and} {Kate
  Compton}. 2017.
\newblock \showarticletitle{Mixed-{Initiative} {Creative} {Interfaces}}. In
  {\em Proceedings of the 2017 {CHI} {Conference} {Extended} {Abstracts} on
  {Human} {Factors} in {Computing} {Systems} - {CHI} {EA} '17}. ACM Press,
  Denver, Colorado, USA, 628--635.
\newblock
\showISBNx{978-1-4503-4656-6}
\showDOI{%
\url{http://dx.doi.org/10.1145/3027063.3027072}}


\bibitem{townley_identity_2009}
{Kimberly~D. Elsbach}. 2009.
\newblock \showarticletitle{Identity affirmation through `signature style': {A}
  study of toy car designers}.
\newblock {\em Human Relations\/} {62}, 7 (July 2009), 1041--1072.
\newblock
\showISSN{0018-7267, 1741-282X}
\showDOI{%
\url{http://dx.doi.org/10.1177/0018726709335538}}


\bibitem{eysenck_creativity_1995}
{Hans~J. Eysenck}. 1995.
\newblock \showarticletitle{Creativity as a {Product} of {Intelligence} and
  {Personality}}.
\newblock In {\em International {Handbook} of {Personality} and
  {Intelligence}}, {Donald~H. Saklofske} {and} {Moshe Zeidner} (Eds.). Springer
  US, Boston, MA, 231--247.
\newblock
\showISBNx{978-1-4757-5571-8}
\showDOI{%
\url{http://dx.doi.org/10.1007/978-1-4757-5571-8_12}}


\bibitem{department_for_digital_culture_media_and_sport_dcms_2018}
{Department for Digital Culture~Media} {and} {Sport}. 2018.
\newblock {\em {DCMS} {Sectors} {Economic} {Estimates} 2017 (provisional):
  {Gross} {Value} {Added}}.
\newblock {T}echnical {R}eport.
\newblock
\showURL{%
\url{https://assets.publishing.service.gov.uk/government/uploads/system/uploads/attachment_data/file/759707/DCMS_Sectors_Economic_Estimates_2017__provisional__GVA.pdf}}


\bibitem{department_for_digital_culture_media_and_sport_dcms_2019}
{Department for Digital Culture~Media} {and} {Sport}. 2019.
\newblock {\em {DCMS} {Sectors} {Economic} {Estimates} 2018: {Employment}}.
\newblock {T}echnical {R}eport. Department for Digital, Culture, Media and
  Sport, London, UK.
\newblock
\showURL{%
\url{https://assets.publishing.service.gov.uk/government/uploads/system/uploads/attachment_data/file/811903/DCMS_Sectors_Economic_Estimates_2018_Employment_report.pdf}}


\bibitem{frich_mapping_2019}
{Jonas Frich}, {Lindsay MacDonald~Vermeulen}, {Christian Remy}, {Michael~Mose
  Biskjaer}, {and} {Peter Dalsgaard}. 2019.
\newblock \showarticletitle{Mapping the {Landscape} of {Creativity} {Support}
  {Tools} in {HCI}}. In {\em Proceedings of the 2019 {CHI} {Conference} on
  {Human} {Factors} in {Computing} {Systems} - {CHI} '19}. ACM Press, Glasgow,
  Scotland Uk, 1--18.
\newblock
\showISBNx{978-1-4503-5970-2}
\showDOI{%
\url{http://dx.doi.org/10.1145/3290605.3300619}}


\bibitem{gabriel_creativity_2016}
{A. Gabriel}, {D. Monticolo}, {M. Camargo}, {and} {M. Bourgault}. 2016.
\newblock \showarticletitle{Creativity support systems: {A} systematic mapping
  study}.
\newblock {\em Thinking Skills and Creativity\/}  {21} (Sept. 2016), 109--122.
\newblock
\showISSN{18711871}
\showDOI{%
\url{http://dx.doi.org/10.1016/j.tsc.2016.05.009}}


\bibitem{gero_metaphoria:_2019}
{Katy~Ilonka Gero} {and} {Lydia~B. Chilton}. 2019.
\newblock \showarticletitle{Metaphoria: {An} {Algorithmic} {Companion} for
  {Metaphor} {Creation}}. In {\em Proceedings of the 2019 {CHI} {Conference} on
  {Human} {Factors} in {Computing} {Systems} - {CHI} '19}. ACM Press, Glasgow,
  Scotland Uk, 1--12.
\newblock
\showISBNx{978-1-4503-5970-2}
\showDOI{%
\url{http://dx.doi.org/10.1145/3290605.3300526}}


\bibitem{madureira_emerging_2017}
{Frederica Gonçalves}, {Eduardo Fermé}, {and} {João~C. Ferreira}. 2017.
\newblock \showarticletitle{Emerging {Opportunities} for {Ambient}
  {Intelligence} in {Creativity} {Support} {Tools}}.
\newblock In {\em Intelligent {Systems} {Design} and {Applications}},
  {Ana~Maria Madureira}, {Ajith Abraham}, {Dorabela Gamboa}, {and} {Paulo
  Novais} (Eds.). Vol. 557. Springer International Publishing, Cham, 640--648.
\newblock
\showISBNx{978-3-319-53479-4 978-3-319-53480-0}
\showDOI{%
\url{http://dx.doi.org/10.1007/978-3-319-53480-0_63}}


\bibitem{google_share_2019}
{{Google}}. 2019.
\newblock Share and engage with the {Design} {Sprint} {Community}.
\newblock   (2019).
\newblock
\showURL{%
\url{https://designsprintkit.withgoogle.com/introduction/overview}}


\bibitem{holmquist_intelligence_2017}
{Lars~Erik Holmquist}. 2017.
\newblock \showarticletitle{Intelligence on tap: artificial intelligence as a
  new design material}.
\newblock {\em interactions\/} {24}, 4 (June 2017), 28--33.
\newblock
\showISSN{10725520}
\showDOI{%
\url{http://dx.doi.org/10.1145/3085571}}


\bibitem{lawrie_could_2019}
{Eleanor Lawrie}. 2019.
\newblock \showarticletitle{Could a computer ever create better art than a
  human?}
\newblock {\em BBC News\/} (April 2019).
\newblock
\showURL{%
\url{https://www.bbc.com/news/business-47700701}}


\bibitem{loi_co-designing_2019}
{Daria Loi}, {Christine~T. Wolf}, {Jeanette~L. Blomberg}, {Raphael Arar}, {and}
  {Margot Brereton}. 2019.
\newblock \showarticletitle{Co-designing {AI} {Futures}: {Integrating} {AI}
  {Ethics}, {Social} {Computing}, and {Design}}. In {\em Companion
  {Publication} of the 2019 on {Designing} {Interactive} {Systems} {Conference}
  2019 {Companion}} {\em ({DIS} '19 {Companion})}. ACM, New York, NY, USA,
  381--384.
\newblock
\showISBNx{978-1-4503-6270-2}
\showDOI{%
\url{http://dx.doi.org/10.1145/3301019.3320000}}
\newblock
\shownote{event-place: San Diego, CA, USA.}


\bibitem{lubart_how_2005}
{Todd Lubart}. 2005.
\newblock \showarticletitle{How can computers be partners in the creative
  process: {Classification} and commentary on the {Special} {Issue}}.
\newblock {\em International Journal of Human-Computer Studies\/} {63}, 4-5
  (Oct. 2005), 365--369.
\newblock
\showISSN{10715819}
\showDOI{%
\url{http://dx.doi.org/10.1016/j.ijhcs.2005.04.002}}


\bibitem{malsattar_designing_2019}
{Nirav Malsattar}, {Tomo Kihara}, {and} {Elisa Giaccardi}. 2019.
\newblock \showarticletitle{Designing and {Prototyping} from the {Perspective}
  of {AI} in the {Wild}}. In {\em Proceedings of the 2019 on {Designing}
  {Interactive} {Systems} {Conference}} {\em ({DIS} '19)}. ACM, New York, NY,
  USA, 1083--1088.
\newblock
\showISBNx{978-1-4503-5850-7}
\showDOI{%
\url{http://dx.doi.org/10.1145/3322276.3322351}}
\newblock
\shownote{event-place: San Diego, CA, USA.}


\bibitem{united_nations_creative_2018}
{United Nations}. 2018.
\newblock {\em Creative {Economy} {Outlook}:{Trends} in international trade in
  creative industries 2002–2015}.
\newblock {T}echnical {R}eport. UNCTAD.
\newblock


\bibitem{sautoy_true_nodate}
{Marcus~du Sautoy}. 2019.
\newblock True {AI} creativity is coming and will reveal the minds of machines.
\newblock   (2019).
\newblock
\showURL{%
\url{https://www.newscientist.com/article/mg24232292/}}


\bibitem{shneiderman_creativity_2007}
{Ben Shneiderman}. 2007.
\newblock \showarticletitle{Creativity support tools: accelerating discovery
  and innovation}.
\newblock {\it Commun. ACM} {50}, 12 (Dec. 2007), 20--32.
\newblock
\showISSN{00010782}
\showDOI{%
\url{http://dx.doi.org/10.1145/1323688.1323689}}


\bibitem{redondo_creativity_2009}
{Ben Shneiderman}. 2009.
\newblock \showarticletitle{Creativity {Support} {Tools}: {A} {Grand}
  {Challenge} for {HCI} {Researchers}}.
\newblock In {\em Engineering the {User} {Interface}}, {Miguel Redondo},
  {Crescencio Bravo}, {and} {Manuel Ortega} (Eds.). Springer London, London,
  1--9.
\newblock
\showISBNx{978-1-84800-135-0 978-1-84800-136-7}
\showDOI{%
\url{http://dx.doi.org/10.1007/978-1-84800-136-7_1}}


\bibitem{sternberg_handbook_1999}
{Robert~J. Sternberg} (Ed.). 1999.
\newblock {\em Handbook of creativity}.
\newblock Cambridge University Press, Cambridge, U.K. ; New York.
\newblock
\showISBNx{978-0-521-57285-9 978-0-521-57604-8}


\bibitem{still_history_nodate}
{Arthur Still} {and} {Mark d'Inverno}.
\newblock \showarticletitle{A {History} of {Creativity} for {Future} {AI}
  {Research}}. In {\em Proceedings of the {Seventh} {International}
  {Conference} on {Computational} {Creativity}, {June} 2016}. 8.
\newblock
\showISBNx{978-2-7466-9155-1}


\bibitem{yang_mapping_2018}
{Qian Yang}, {Nikola Banovic}, {and} {John Zimmerman}. 2018a.
\newblock \showarticletitle{Mapping {Machine} {Learning} {Advances} from {HCI}
  {Research} to {Reveal} {Starting} {Places} for {Design} {Innovation}}. In
  {\em Proceedings of the 2018 {CHI} {Conference} on {Human} {Factors} in
  {Computing} {Systems} - {CHI} '18}. ACM Press, Montreal QC, Canada, 1--11.
\newblock
\showISBNx{978-1-4503-5620-6}
\showDOI{%
\url{http://dx.doi.org/10.1145/3173574.3173704}}


\bibitem{yang_investigating_2018}
{Qian Yang}, {Alex Scuito}, {John Zimmerman}, {Jodi Forlizzi}, {and} {Aaron
  Steinfeld}. 2018b.
\newblock \showarticletitle{Investigating {How} {Experienced} {UX} {Designers}
  {Effectively} {Work} with {Machine} {Learning}}. In {\em Proceedings of the
  2018 {Designing} {Interactive} {Systems} {Conference}} {\em ({DIS} '18)}.
  ACM, New York, NY, USA, 585--596.
\newblock
\showISBNx{978-1-4503-5198-0}
\showDOI{%
\url{http://dx.doi.org/10.1145/3196709.3196730}}
\newblock
\shownote{event-place: Hong Kong, China.}


\end{thebibliography}

\end{document}